\documentclass[letterpaper, 10 pt, conference]{ieeeconf}  

\IEEEoverridecommandlockouts                              

\usepackage{graphics} 
\usepackage{epsfig} 
\usepackage{amsmath} 
\usepackage{amssymb}  
\usepackage{bm}
\usepackage{lipsum}
\usepackage{cite}
\usepackage{multicol}
\usepackage{multirow}
\usepackage{algpseudocode}

\usepackage{acro}
\usepackage{styles/acro}
\usepackage{styles/acro-CSE}
\usepackage{styles/acro-ST}

\usepackage{xcolor}
\usepackage{comment}
\usepackage[ruled, vlined]{algorithm2e}
\usepackage{tabularx}
\usepackage{graphicx}

\usepackage{xcolor}
\usepackage[ruled, vlined]{algorithm2e}

\SetCommentSty{mycommfont}

\SetKwInput{Kwinit}{\textcolor{blue}{Initialization}}            
\SetKwInput{Kwinitend}{\textcolor{blue}{end Initialization}}     
\SetKwInput{Kwtrain}{\textcolor{blue}{Training}}                 
\SetKwInput{Kwtrainend}{\textcolor{blue}{end Training}}          
\SetKwFunction{KwFn}{\textbf{BestAction}($Fk_{net}$,  $a_p$)}    

\usepackage{tikz}
\usetikzlibrary{fit,calc}

\colorlet{pink}{red!40}
\colorlet{cyan}{cyan!60}

\usepackage{hyperref}

\newcommand\identity[2][\empty \empty]{
  \ifthenelse{\equal{#2}{\empty}}
    {\vI}
    {\ifthenelse{\equal{#1}{\empty}}
        {\vI_{#2}}
        {\vI_{#1\!\times\!#2}}
    }
}
\newcommand\zero[2][\empty \empty]{
  \ifthenelse{\equal{#2}{\empty}}
    {\mathbf{0}}
    {\ifthenelse{\equal{#1}{\empty}}
        {\mathbf{0}_{#2}}
        {\mathbf{0}_{#1\!\times\!#2}}
    }
}

\newcommand\Real[2][\empty \empty]{
  \ifthenelse{\equal{#2}{\empty}}
    {\mathbb{R}}
    {\ifthenelse{\equal{#1}{\empty}}
        {\mathbb{R}^{#2}}
        {\mathbb{R}^{#1\!\times\!#2}}
    }
}
\newcommand\inReal[2][\empty \empty]{
  \ifthenelse{\equal{#2}{\empty}}
    {\in \Real{}}
    {\ifthenelse{\equal{#1}{\empty}}
        {\in \Real{#2}}
        {\in \Real[#1]{#2}}
    }
}

\newcommand{\vI}{\mathbf I}

\title{\LARGE \bf
{Autonomous Control of Redundant Hydraulic Manipulator
Using Reinforcement Learning with Action Feedback}
}

\author{Rohit Dhakate$^{1}$,  {Christian Brommer}$^{1}$, {Christoph Böhm}$^{1}$, {Harald Gietler}$^{2}$,  {Stephan Weiss}$^{1}$, and {Jan Steinbrener}$^{1}$
\thanks{$^{1}$Rohit Dhakate, Christian Brommer, Christoph Böhm, Stephan Weiss and Jan Steinbrener
        are with the Department of Smart Systems Technologies
        in the Control of Networked Systems Group,
        University of Klagenfurt, 9020 Klagenfurt, Austria
        {\tt\small \{rohit.dhakate, christian.brommer, christoph.boehm, stephan.weiss, jan.steinbrener\}@ieee.org}
        \newline
        $^{2}$Harald Gietler is with the Department of Smart Systems Technologies in the Sensors and Actuators Group,  University of Klagenfurt, 9020 Klagenfurt, Austria
        {\tt\small \{harald.gietler\}@aau.at}
        }
        \thanks{{\textbf{Pre-print version, accepted June/2022, DOI follows ASAP~\copyright IEEE.}}}%
}

\begin{document}

\setlength{\abovedisplayskip}{3pt}
\setlength{\belowdisplayskip}{3pt}

\maketitle
\thispagestyle{empty}
\pagestyle{empty}

\begin{abstract}\label{sec:abstract}
This article presents an entirely data-driven approach for autonomous control of redundant manipulators with hydraulic actuation. 
The approach only requires minimal
system information, which is inherited from a simulation model.
The non-linear hydraulic actuation dynamics are modeled using actuator networks from the data gathered during the manual operation of the manipulator to effectively emulate the real system in a simulation environment. A neural network control
policy for autonomous control, based on end-effector (EE)
position tracking is then learned using Reinforcement Learning
(RL) with Ornstein–Uhlenbeck process noise (OUNoise) for efficient exploration.
The RL agent also receives feedback based on supervised
learning of the forward kinematics which facilitates selecting the best suitable action from exploration. The control policy
directly provides the joint variables as outputs based on
provided target EE position while taking into account the
system dynamics. The joint variables are then mapped to the
hydraulic valve commands, which are then fed to the system
without further modifications. The proposed approach is
implemented on a scaled hydraulic forwarder crane with three
revolute and one prismatic joint to track the desired position of the EE in 3-Dimensional (3D) space. With the emulated dynamics
and extensive learning in simulation, the results demonstrate
the feasibility of deploying the learned controller directly on the
real system.
\end{abstract}

\section{Introduction}\label{sec:introduction}
Hydraulic cranes are versatile heavy-duty manipulators that are omnipresent in construction, mining, agriculture, or forestry for lifting and transporting heavy objects. Automation by sensor retrofitting of these manipulators tackles not only challenging and dull, dangerous, dirty (DDD) tasks concerning the handling of raw materials but also brings economic benefits by increased productivity, and effortless system upgrades according to the desired functionality. 

With the proposed approach, we are addressing the forest log transportation use case. The manipulator repeatedly performs a monotonous pick-and-place operation to collect and redistribute logs prepared by the harvester. Forwarder cranes mainly remain manually operated, despite continuous widespread automation in the industry. Manual operation of such manipulators can be both mentally and physically exhausting, when producing constant, smooth and jerk free motion with joystick, since it requires complex coordination of several hydraulic cylinders \cite{la2019we}.  Early automatic and semi-automatic solutions were presented by \cite{HERA20082306} using analytical methods.
\begin{figure}[!t]
    \includegraphics[width=8.5cm, height=6cm]{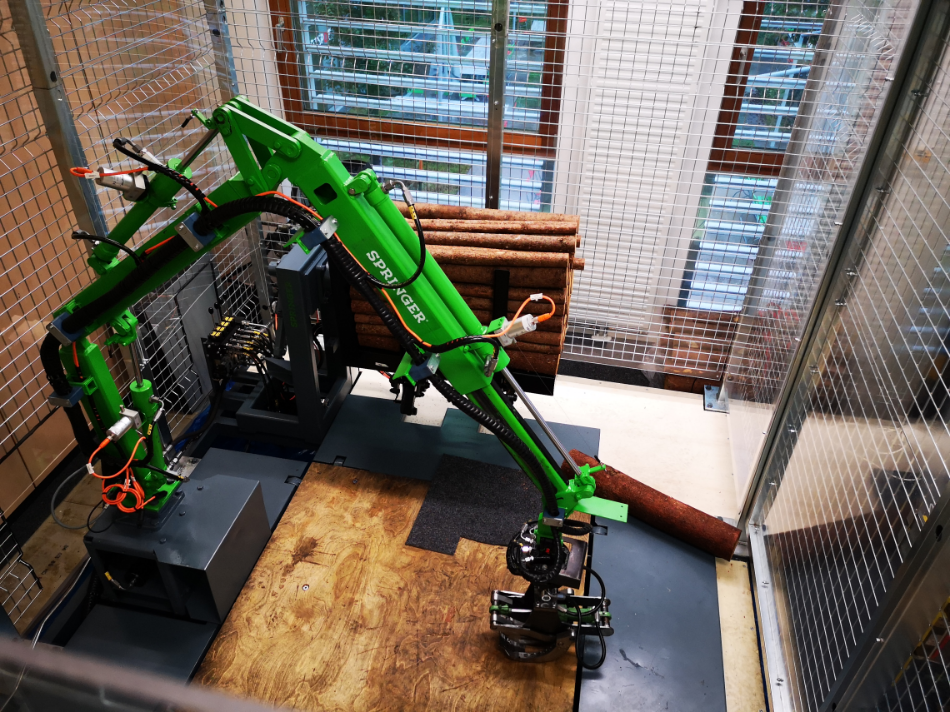}
    \centering
    \caption{AutoLOG manipulator (1:5 scaled forest forwarder crane): Test-bed for our RL-based controller and manipulation tasks.}
    \vspace{-10pt}
    \label{fig:manipulator}
\end{figure}

The barriers in automation of the forest industry can be traced not only towards complex and dynamic environments but also the complexity and variants of the manipulators depending on the manufacturer. In \cite{la2019we} the authors argue that the automation of the entire forwarding operation is complex as numerous tasks such as log recognition, log grasping point detection/selection and pick-and-place operations are involved. However, the authors conclude that the motion patterns of the manipulator's joints are, as expected, highly repetitive and can be automated using analytical methods. However, for analytical methods, an accurate system and environment model is of utmost necessity to achieve desired results, which could take a lot of effort and time given the complexity and redundant nature of the manipulator.

Recent advancements in reinforcement learning not only demonstrated their applications in video-games and simulations but also enabled physical robots to learn complex skills and perform operations in real-world environments. In robot manipulation, reinforcement learning is being extensively used to develop intelligent systems that only require minimal to no system and environment information.
\newpage
\subsection{Related Work}

Autonomous control for forest cranes has been extensively researched for the last two decades. In \cite{la2015model}, the authors modeled the system dynamics using differential equations and applied non-linear control laws, and then performed a calibration and control tuning. While\cite{7364729} also focuses on the aspect of forest crane automation, in addition to compensation for actuator nonlinearities, their main focus is on automating only the base joint (slewing motion). Until recently, all the work done towards automating forest cranes relied on model-based control. 
Current advancements in artificial intelligence (AI) brought substantial simplifications and advantages in tackling complex systems and problems. Within AI, RL algorithms that can be developed in a model-free domain have attracted several researchers and drove the field of automating heavy machinery with the use of AI.

Several RL algorithms have been proposed to solve dynamic physical models in recent years. Among which model-free algorithms gained keen interest due to their nature of generalizing a solution to a category of a problem.
In model-free methods, Q-learning based algorithms such as Deep Q-Network(DQN)\cite{Mnih2015}, Quantile Regression DQN (QR-DQN)\cite{dabney2018distributional}, learns the action-value function $Q(s, a)$ which is the expected value (cumulative discounted reward) of doing an action $a$ in state $s$ and then following the optimal policy, which is deterministic.
Whereas Policy optimization-based algorithms such as Policy gradients , Advantage Actor-Critic (A2C)/ Asynchronous Advantage Actor-Critic (A3C)\cite{mnih2016asynchronous}, Proximal Policy Optimization (PPO)\cite{schulman2017proximal}, and  Trust Region Policy Optimization (TRPO)\cite{pmlr-v37-schulman15}, the agent learns directly the policy function that maps state to action. The policy is determined without using a value function.
In recent years the application of RL for complex manipulation tasks has being carried out by several researchers. In \cite{egli2020towards} the authors implemented a TRPO algorithm for automating a hydraulic excavator. The learned control policy is validated by deploying it on the actual excavator. However, they do not control the base joint which limits the motion in 2D. The authors in \cite{andersson2021reinforcement} use PPO for learning a control policy, along with curriculum learning for grasping tasks. An energy optimization goal is also added in the reward function. However, the validation of the learned policy is conducted on the same simulation platform on which it had been trained.
\subsection{Contribution}
To the best of our knowledge, we present the first work for automation of a real forestry crane with artificial intelligence. Our work investigates the feasibility of applying an actuator-space control policy learned in simulation on a real-world, 4 degrees of freedom, kinematically redundant forestry crane manipulator. The learned control policy maps task-space goals directly to actuator-space commands by providing the target's cartesian position.

We propose a generalized framework for autonomous control of redundant manipulators with highly non-linear hydraulic actuation. The main contributions of the proposed work are listed below,   
\begin{itemize}
    \item Fully data driven approach for position tracking controller of redundant hydraulic manipulator, with minimal system information, negating the need for analytical formulation of forward and inverse kinematics, which is a highly complex task with non-standard manipulators and is subject to change with manipulator models.
    \item Emulated hydraulic actuation dynamics to precisely map from cylinder displacement to joint angles and vice-versa, eliminating the need for formulating the cylinder-joint mapping using geometry. 
    \item Improvement on baseline RL controller, with feedback to predicted actions from forward kinematics network using supervised learning, which directly outputs valve commands for the required target EE position.
    \item A Sim-2-Real deployment of simulation learnt control policy onto real manipulator directly without any adaptation. To the best of our knowledge, this is the first time a Sim-2-Real transfer of RL control policy is deployed on a heavy duty manipulator for 3D position tracking in real-world. The controller performs well in tracking circular and helical trajectories both in simulation and real-world experiments.

\end{itemize}
\section{SYSTEM DESCRIPTION}
\begin{figure}[hbt!]
\includegraphics[width=6cm, height=5.6cm]{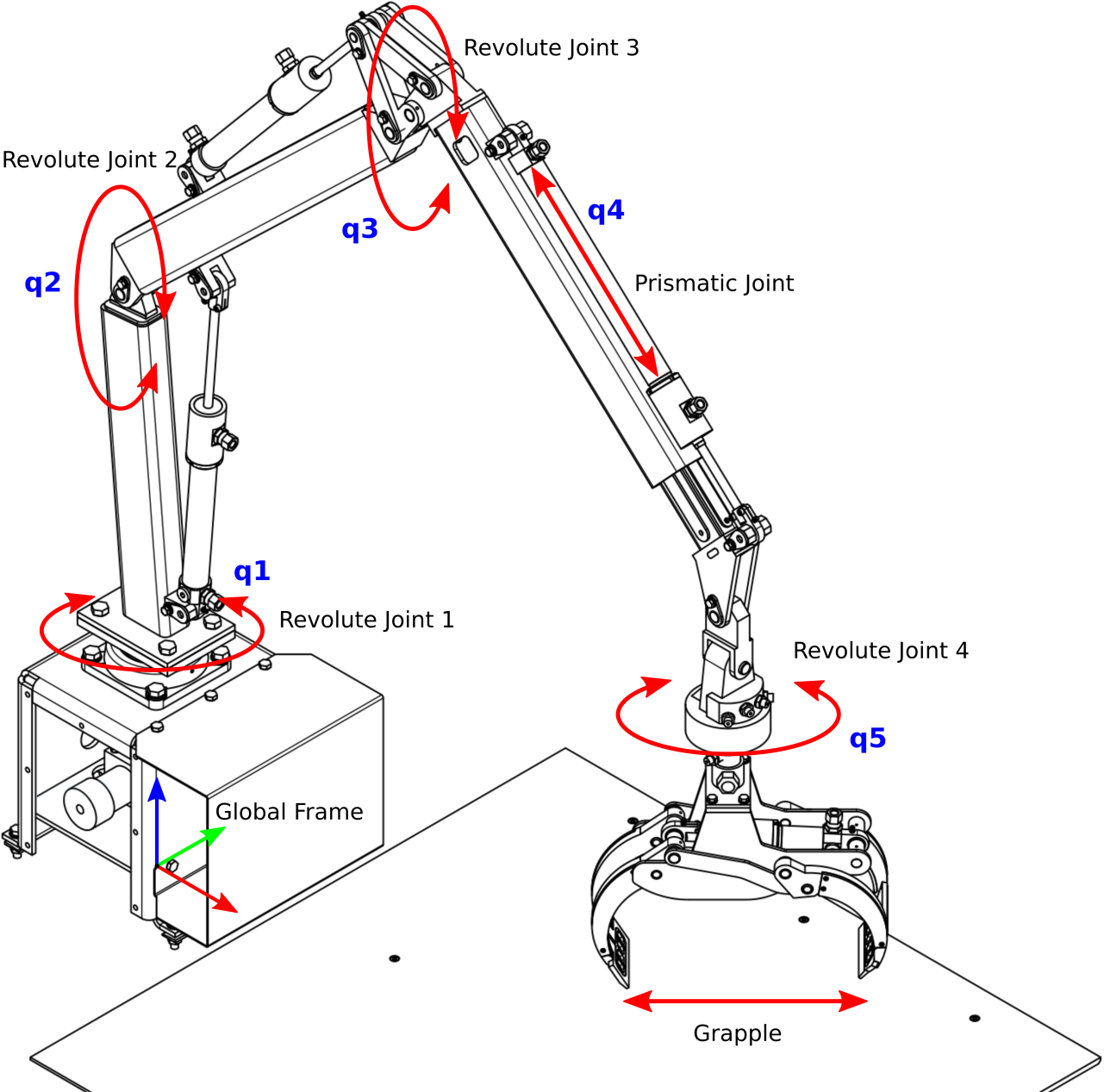}
\centering
\caption{Manipulator description: The figure displays the kinematic configuration of the manipulator. The manipulator has 4 revolute and  1 prismatic joint. All the joints in addition to the grapple are actuated using hydraulic cylinders.}
\label{fig:manipulator_description}
\end{figure}

The Autonomous Log Ordering through Robotic Grasping (AutoLOG) manipulator, which is a 1:5 scaled-down model of an actual forest forwarder crane, is used as a test-bed for autonomous manipulation tasks (see Fig~\ref{fig:manipulator}). The manipulator is powered using hydraulic cylinders for its joint motions. With controllable 5 degrees of freedom, the EE can be controlled for its 3D position and yaw angle, making the manipulator redundant in nature. However, for our learning-based control task, we omit the yaw component and only focus on the 3D position of the EE. The yaw parameter of the system is application dependent, such as aligning the yaw with respect to the log orientation for pick-and-place tasks. A description of manipulator configuration is shown in Fig~\ref{fig:manipulator_description}.

\begin{enumerate}
    \item \textit{Joint Angles / Displacement Measurements}: For forest crane actuators, direct access to the inputs and outputs is not always available. Hence we retrofit our manipulator with exteroceptive sensors. The manipulator comprises three revolute and one prismatic joint. Just before the revolute joint 4, we have 2 orthogonal under-actuated joints, which cause the gripper to sway freely in a 3D space. The joint states for revolute joints 2, 3, and prismatic joint are obtained by mapping the cylinder displacements to joint angles. We use Waycon SX50 draw-wire sensors to measure the cylinder displacements with a measurement error of 0.0002 mm over a displacement of 1250 mm. For revolute joint 1, we use a retrofitted inductance-based angular position sensor which provides absolute angle measurement with a maximum measurement error of 0.8 degrees  \cite{gietler2020scalable}.
    \item \textit{Electric Control Valves}: For autonomous control of the manipulator, the hydraulic proportional valves (electro-hydraulic) are controlled using a Pulse Width Modulation (PWM) control which changes the fluid flow in cylinders according to required joint values.

    \item \textit{Requirements for Approach}: Our proposed method requires minimal system information. Table \ref{table: controller I/O} lists the inputs and outputs of our proposed approach.

    \begin{table}[hbt!]
    \centering
    \caption{RL Agent -  Inputs and Outputs}
        \begin{tabular}{|c|l|l|}
        \hline
            Parameters              &Inputs                     &Output     \\
            \hline
            \hline
            Joint values            &$q_{t}$                    &$q_{t+1}$  \\
            \hline
            Current EE position     &$X_{t}$                    &            \\
            \hline
            Target EE position      &$X_{t+1}$                  &           \\
        \hline
        \end{tabular}
    \label{table: controller I/O}
    \end{table}
\end{enumerate}
\section{SIMULATION FRAMEWORK}
\begin{figure}[ht]
            \includegraphics[width=7cm, height=6cm]{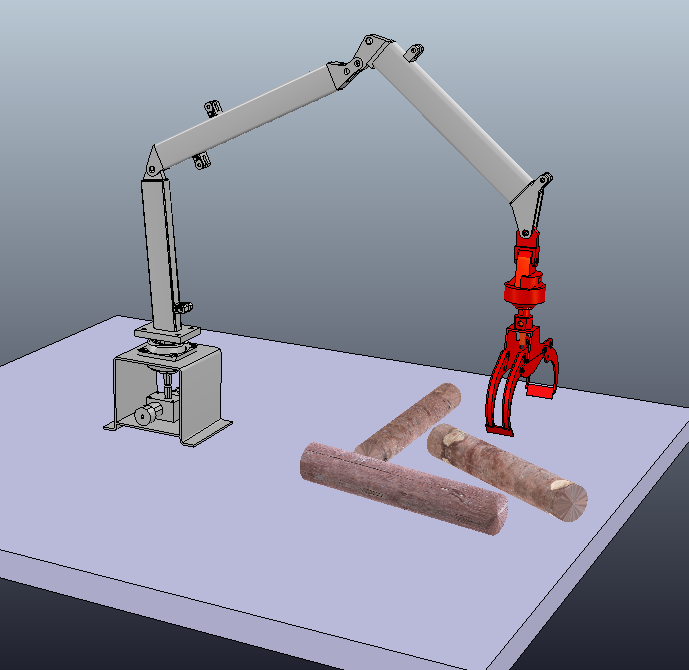}
            \centering
            \caption{Simulation model in CoppeliaSim. The gripper (red object) is detached for training since the effects of gripper sway is out of the scope of the proposed approach.}
            \label{fig:sim_model}
            \end{figure}

We use CoppeliaSim (formerly V-REP) \cite{rohmer2013v} as our simulation framework to train the RL agent. CoppeliaSim provides a wide range of functionalities and supports multiple physics engines including Bullet \cite{coumans2013bullet}, ODE \cite{smith2007open}, Vortex \cite{cm2020vortex} and Newton \cite{jerez2008newton}. The simulation scene is generated using a Computer Aided Design (CAD) model of the manipulator. The scene is dynamically enabled using Bullet 2.78 physics engine to render our simulation. The simulator provides a kinematics calculation module to compute forward and inverse kinematics of the manipulator chain, however we only use the position information of the scene objects (joints and end-effector) for our observations. Observations from the simulator can be considered as measurement from our retrofitted sensors on the real system. To control the manipulator we use a python remote API Client.
\newline
The simulation model is shown in Fig~\ref{fig:sim_model}. The gripper (red~object) is detached for simulations. Thus our simulation setup does not have the cylinder displacements as control inputs. Instead, the joint variables are provided directly to the simulator. However, the resulting joint variables from the learned controller are converted to cylinder displacements using the actuator network.

\section{METHODS}
    \subsection{Overview}
\begin{figure}[!htpb]
        \includegraphics[width=0.35\textwidth]{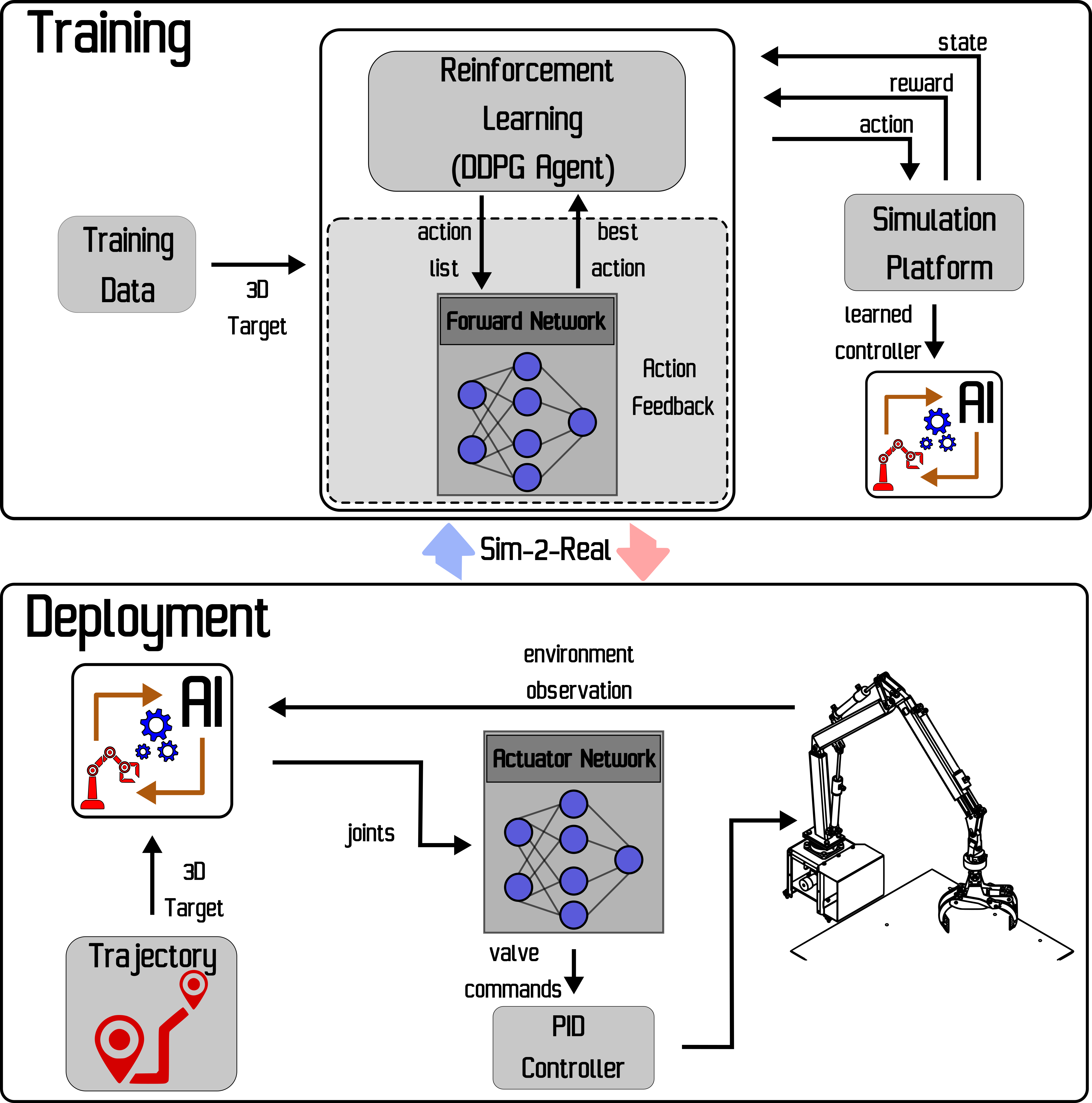}
        \centering
        \caption{RL control architecture: The image shows an architecture overview of our proposed approach. The training is done completely offline on a simulation platform, it shows the interaction between forward network, RL agent and the simulation platform. Sim-2-Real transfer of the trained controller is validated by directly deploying it on the physical system.}
        \label{fig:system_architecture}
    \end{figure}

    Fig~\ref{fig:system_architecture} shows an overview of our approach. We train an actuator and a forward network using supervised learning. The actuator network incorporates the non-linear dynamics involved in the hydraulic actuation and is trained to map cylinder displacement to joint variables and vice-versa. The forward network is a mapping from joint space to operation space of the manipulator. The RL agent (DDPG) is then trained in the simulation to reach a target 3D position from a random initial joint configuration. The trained RL agent is first evaluated for a trajectory tracking task in simulation and then is deployed on the real manipulator for final validation.
    
    \subsection{Network Modelling}
    The two supervised learning networks facilitate our approach of learning based control.

    \begin{enumerate}
        \item Actuator Network: The actuator network performs a bi-directional mapping between cylinder displacements and joint variables. Our RL agent outputs joint variables for the target goal, whereas low-level manipulator control takes valve commands (cylinder displacements) as control inputs.
        \item Forward Network: The forward network takes current joint variables as inputs and returns the 3D position of EE.
    \end{enumerate}
    
    \subsection{Data Collection}
    \begin{enumerate}
        \item Actuator Network Data: For the actuator network, we collected input-output data from the real system during manual operation. We recorded the cylinder displacements using the retrofitted draw-wire sensors, and a motion capture system is used to measure the respective angles, since our system does not have an alternative for direct angle measurement for revolute joints 2 and 3. The cylinder control inputs were provided using a remote control designed for the manipulator. The data is collected with different cylinder velocities to capture the hydraulic actuation dynamics effectively. The collected data is believed to be incorporating all the non-linear dynamics involved in the mapping between cylinders and respective angles, see Fig~\ref{fig:train_data_actuator}.  
            \begin{figure}[!hbt]
            \includegraphics[width=7.5cm, height=5cm]{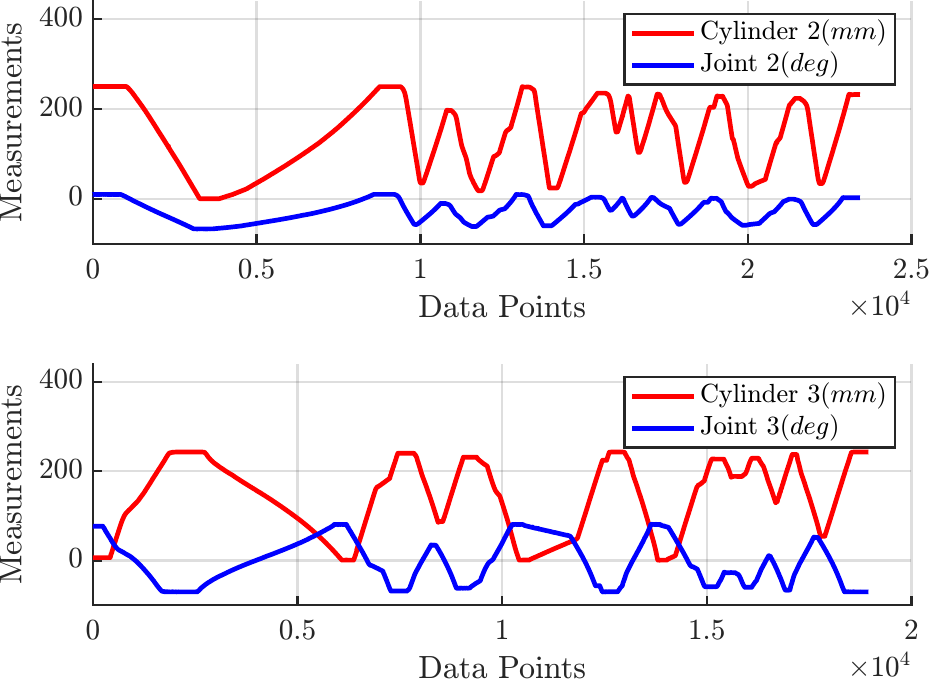}
            \centering
            \caption{Real data to train the actuator model is gathered from the physical manipulator. The figure shows the joint angles w.r.t. the cylinder displacement. The cylinder displacements are measured using a draw-wire sensor and corresponding joint angles are recorded using motion capture system.}
            \label{fig:train_data_actuator}
            \end{figure}
        \item Forward Network Data: To train the forward network, we acquired the joint variables and EE position data autonomously by setting a random joint configuration for each data point and recording the EE position using motion capture, as shown in Fig~\ref{fig:train_data_fk}. The collected data also gave an insight into the manipulator work-space. 
            \begin{figure}[!hbt]
            \includegraphics[width=7.5cm, height=5cm]{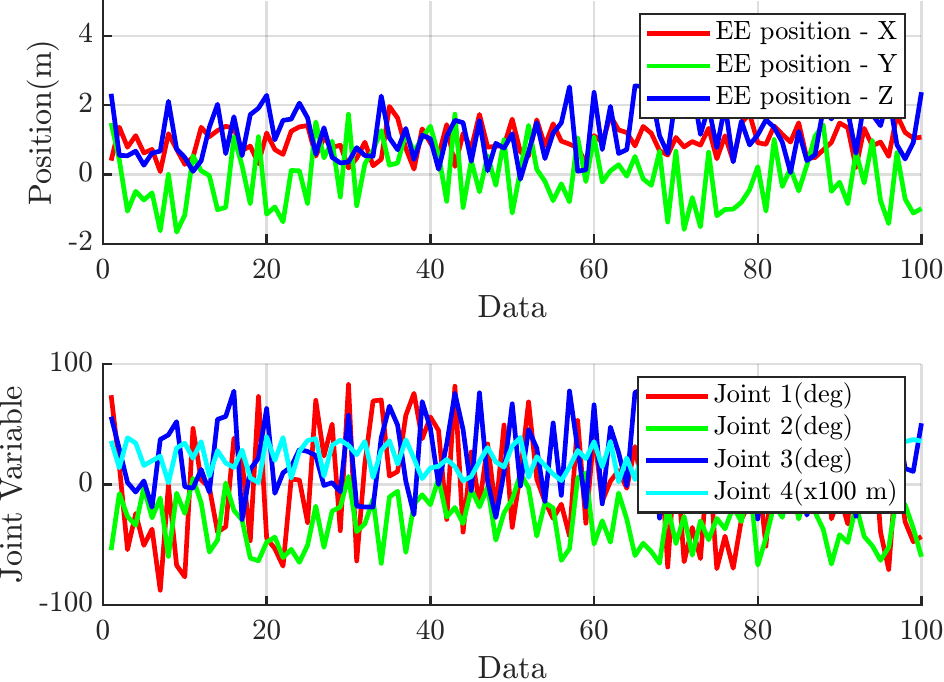}
            \centering
            \caption{The image displays the real data recorded for forward kinematics in an autonomous fashion, to train our forward network. For each data-point random joint configuration is set and corresponding EE position is recorded using motion capture system. The sampled random configurations covers the full range of cylinder displacements.}
            \vspace{-15pt}
            \label{fig:train_data_fk}
            \end{figure}
    \end{enumerate}

    \subsection{Network Training}
    \begin{enumerate}
    \item Actuator Network Training: We train separate networks for each joint-cylinder mapping. Actuator network-2 maps joint2-cylinder2, while actuator network-3 maps joint3-cylinder3.
    The actuator network-2 is trained using a simple multi-layer perceptron (MLP) with 3 hidden layers (with 256-128-128 hidden units) and non-linear rectified linear unit (ReLU) activation. We used Adam optimizer with a learning rate of 1e-4. The model predicts the cylinder position for a given joint angle. Whereas the actuator network-3 uses an MLP with only 2 hidden layers(with 128-128 hidden units). 
    Fig~\ref{fig:act_val_j2} and~\ref{fig:act_val_j3} shows the validation of trained actuator networks.
    \begin{figure}[!htpb]
        \includegraphics[width=7.5cm, height=5cm]{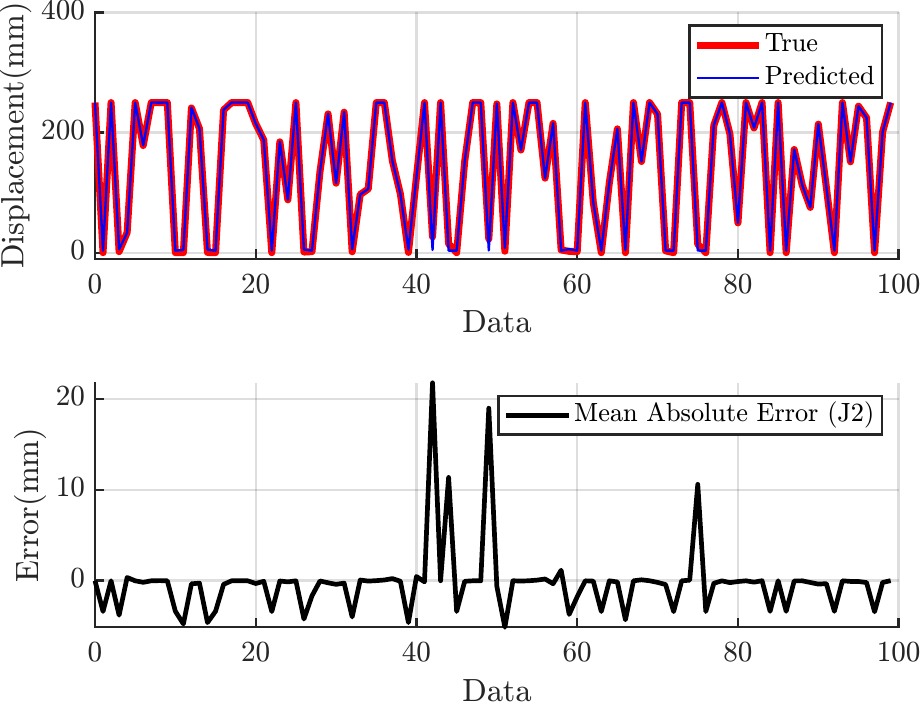}
        \centering
        \caption{The figure shows validation results of trained actuator network for joint 2. Excluding error spikes at few instances the network precisely learned the cylinder-joint mapping.}
        \label{fig:act_val_j2}
    \end{figure}
    \begin{figure}[!htpb]
        \includegraphics[width=7.5cm, height=5cm]{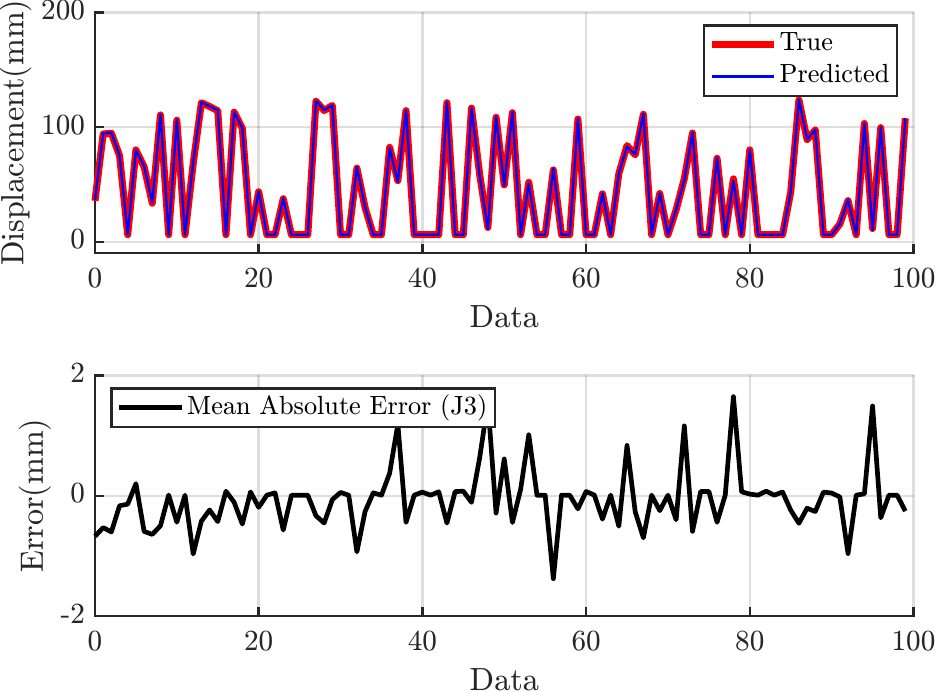}
        \centering
        \caption{Validation results of the trained actuator network for joint 3 are shown in this figure.}
        \vspace{-5pt}
        \label{fig:act_val_j3}
    \end{figure}
    
    \item Forward Network Training: Our forward network is a multi-input-multi-output (MIMO) mapping from joint variables to EE position. The network is trained using an MLP with only 2 hidden layers (with 256-128 hidden units). Despite training the network on only 500 data points, Fig~\ref{fig:fk_validation} shows that the generalization is very accurate with a maximum prediction error of only (0.0159, 0.0205, 0.0136)m in x, y, and z, respectively.
    \begin{figure}[ht]
        \includegraphics[width=7.5cm, height=5cm]{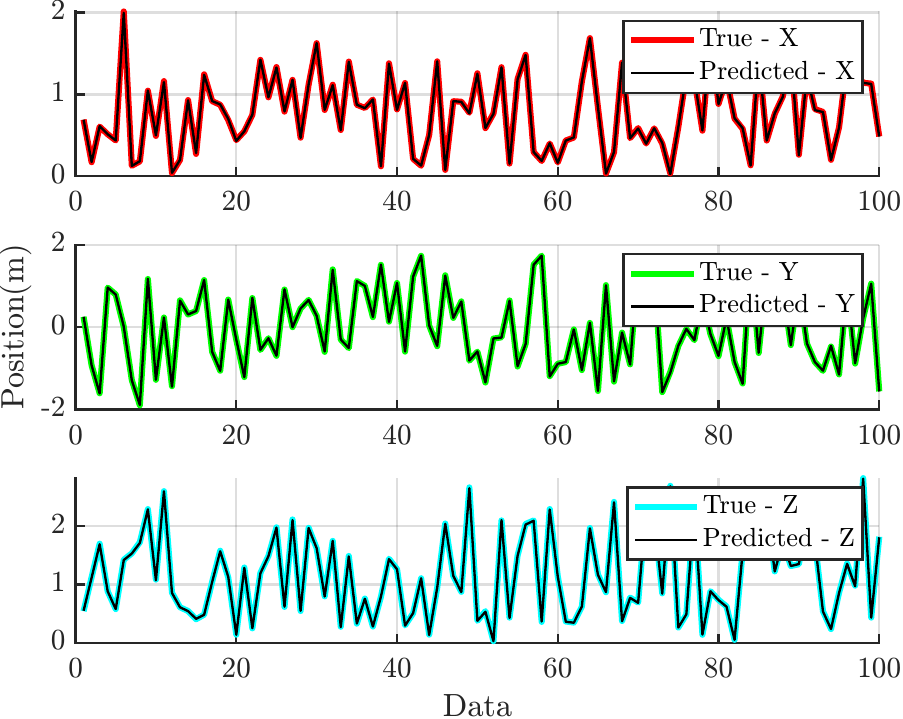}
        \centering
        \caption{Figure shows evaluation of the forward network. With a multi-input structure involving 4 joint variables, the network trained very efficiently to return 3D position of the EE.}
        \vspace{-15pt}
        \label{fig:fk_validation}
    \end{figure}
    
    \end{enumerate}
 
    \subsection{Reinforcement Learning Controller}
    Our proposed learning-based controller uses RL to synthesize a model-free task-space position tracking controller. The RL controller learns the inverse kinematics of the manipulator, which cannot be formulated analytically without any optimization objectives due to the redundant nature of the manipulator. 
    
\begin{figure}[ht]
        \includegraphics[width=4.5cm, height=2.8cm]{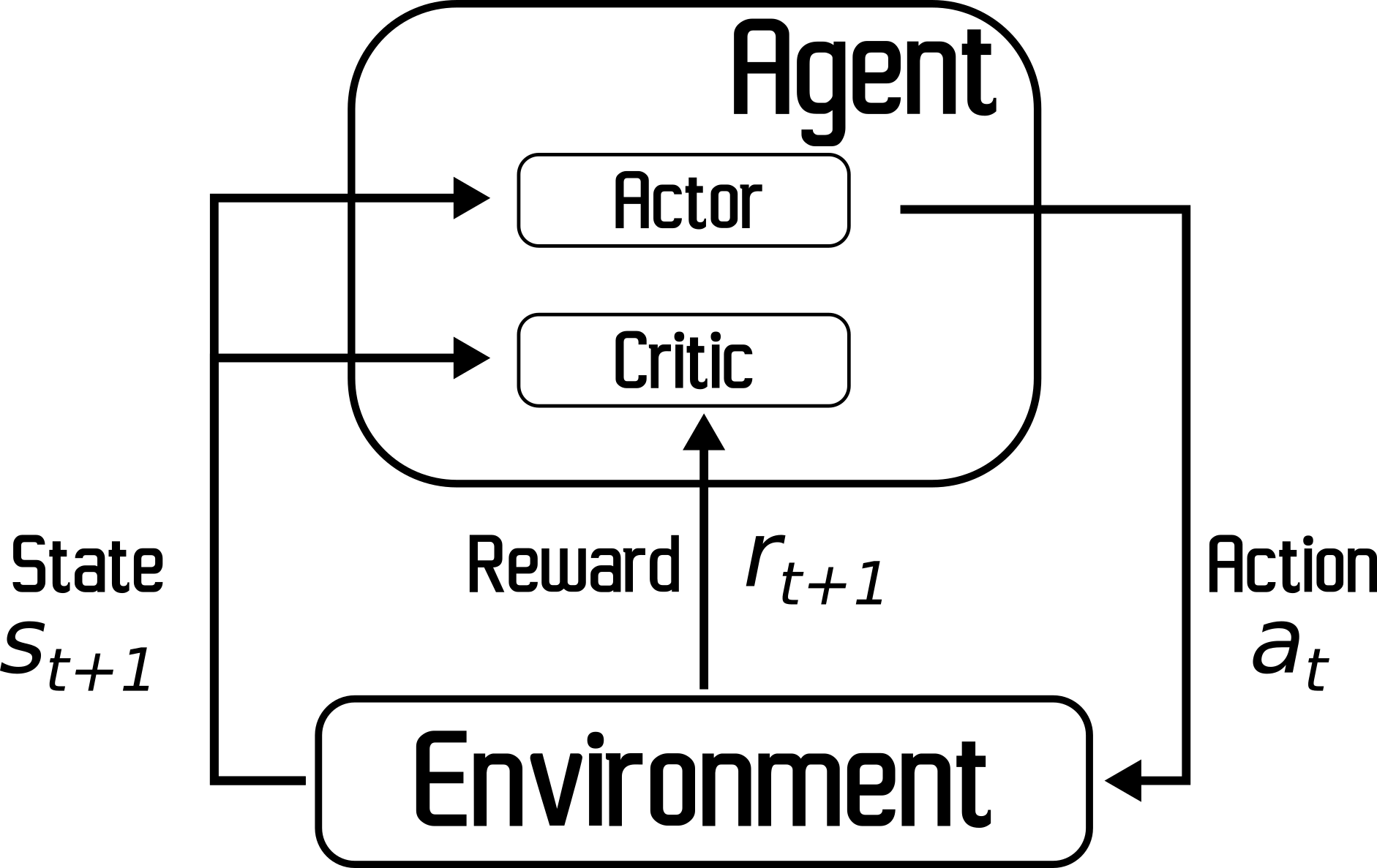}
        \centering
        \caption{A simple architecture of actor-critic method of RL approach is described. It shows the main operation of any RL based algorithm, with $state$, $action$, $reward$, $agent$ and $environment$ being the main components of an RL algorithm.}
        \vspace{-15pt}
        \label{fig:simple_rl}
    \end{figure}

    Reinforcement Learning Preliminaries:
    
    We formalized our RL problem as a Markov decision process (MDP), which is a discrete-time stochastic control process. We use MDP, which provides a mathematical framework for predicting outcomes where the environment is fully observable. The MDP is characterized by,
    
    \begin{itemize}
        \item state ($s$)   : state of the agent in the environment 
        \item action ($a$)  : predicted/ conducted action by the agent 
        \item reward ($r$)  : a scalar valued reward based on performed action and achieved state
        \item policy ($\pi(s|a)$): decision making function of state-action pair
    \end{itemize}
    A simple actor-critic architecture is shown in Fig~\ref{fig:simple_rl}.
    
    At a given discrete time step t, the state of the system is given by, $s_{t} \in S$. The agent makes an observation of the environment $o_{t} \in O$. Performing an action $a_{t} \in A$ according to the policy distribution $\pi(a|s)$, the agent receives an immediate scalar reward $r_{t}(s_{t}, a_{t})$ according to the specified reward function $R(s, a)$ providing an updated state $s^{\mathrm{'}}_{\mathrm{t+1}} \in S$. The goal of RL algorithms is to find the optimal policy $\pi^\ast(a|s)$, such that the agent takes the optimal action at any given state in order to maximize the expected return. Here, the deep RL approach involves parameterizing the policy $\pi$ as a neural network $\pi(\theta)$ with parameters $\theta \in \Theta$. The resulting policy approximator outputs a vector of actuator-space control signals at each time step.

    We use DDPG\cite{lillicrap2015continuous} because it combines both Q-learning and policy optimization approaches. DDPG has an actor-critic architecture, where the critic network determines the Q value, and the actor network determines the actions to be taken. The actor network in DDPG simply uses the negative average Q value generated by the critic model as a loss and learns to generate actions to maximize the Q value in each state. An experience replay buffer stores all the experiences and draws a batch to train the networks. To the DDPG baseline, we added feedback to the predicted actions using our forward network for efficient exploration. Using the current policy, we predict a specified number of actions, which is then fed to the forward network to find the best actions based on the norm distance between EE position from predicted actions and the target position. The selected action is then used to perform a ($s_{t}, a_{t}, r_{t}, s^{\mathrm{'}}_{\mathrm{t+1}}$) step to get the next state $s^{\mathrm{'}}_{\mathrm{t+1}}$.
    The contents of our system state and actions are described in Table \ref{tabel: DDPG components}.
    \begin{table}[hbt!]
    \centering
    \caption{DDPG Algorithm Components}
        \begin{tabular}{|c|c|c|}
        \hline
        Parameters & Contents & Dimension \\ 
        \hline
        \hline
        \multirow{3}{*}{State}      & \multirow{1}{*}{Observation: Joint variables} 
                                    & \multirow{1}{*}{4x1}  \\ \cline{2-3}
                                    
                                    & \multirow{1}{*}{Achieved goal: Current EE} 
                                    & \multirow{1}{*}{3x1}  \\ \cline{2-3}
                                    
                                    & \multirow{1}{*}{Desired goal: Target EE}  
                                    & \multirow{1}{*}{3x1}  \\ \cline{2-3}
        \hline
        \hline
        \multirow{1}{*}{Actions}   & \multirow{1}{*}{Joint Variables: $[J_1, J_2, J_3, J_4]$} & \multirow{1}{*}{4x1}  \\
        \hline
        \end{tabular}
    \label{tabel: DDPG components}
    \end{table}
    
    We give a constant reward $r^{{step}}_{{t}}$ for each time-step which improves the learning performance. A distance reward $r^{{dist}}_{{t}}$ which helps to learn the reaching task is given based on the norm distance between current and target EE position. We also add a joint limit avoidance reward $r^{{jlim}}_{{t}}$ which discourages the agent from learning infeasible joint configurations. A complete episode reward is the sum of all the aforementioned reward functions.
    \\
    
    Our reward function is defined as follows, 
    \begin{equation} \label{eu_eqn}
    r_{t} = r^{\mathrm{step}}_{\mathrm{t}} + r^{\mathrm{dist}}_{\mathrm{t}} + r^{\mathrm{jlim}}_{\mathrm{t}}
    \end{equation}
    \newpage
    where,
    \\
    
    $\begin{aligned}
        r^{{step}}_{{t}}     &= 0.001\\
        r^{{dist}}_{{t}}     &= -(||x_{t+1} - x_{t}||_2) + 0.002\\
        r^{{jlim}}_{{t}}     &=
         \begin{cases}
        -0.0005,  & \text{if, $j > j_{max}$ or $j < j_{min}$} \\
        0, & \text{if,  $j_{min} \geq j \leq j_{max}$}\\
         \end{cases}
    \end{aligned}$\\

    We start the simulation with random initial joint configuration, and our RL agent acquires observations from the simulation environment which forms our system state. The actor network then generates random actions based on the current state and exploration noise. These actions (joint variables as control inputs) are then filtered using the forward network to select the best action, which is then carried out in a simulation step. The simulation is carried out at 100hz, the same rate our real system is operated.
    
    The DDPG algorithm is claimed to be sensitive to hyperparameters, which we observed during tuning of the hyperparameters. In\cite{DBLP:journals/corr/abs-1709-06560}, it is shown that DDPG with tuned hyperparameters outperforms several other policy optimization algorithms in stable environments. We modified the hyperparameters from the stable baseline parameters to suit our training environment. Table~\ref{table: hyperparameters} shows the hyperparameters used for our system. 
        \begin{table}[!hbt]
    \centering
    \caption{Algorithm Hyperparameters}
        \begin{tabular}{|c|c|l|}
        \hline
            Parameters                  & Variable              &Values \\
            \hline
            \hline
            Number of episodes          &$n_{episodes}$          &$1500$  \\
            \hline
            Number of steps             &$n_{steps}$            &$1000$  \\
            \hline
            Buffer size                 &$n_{buffer}$             &$1e+06$ \\
            \hline
            Batch size                  &$n_{batch}$              &$1024$  \\
            \hline
            Discount factor             &$\gamma$            &$0.99$  \\
            \hline
            Soft target update          &$\tau$              &$1e-03$  \\
            \hline
            Actor learning rate         &$lr_{ac}$                &$1e-03$  \\
            \hline
            Critic learning rate        &$lr_{cr} $               &$1e-03$  \\
            \hline
            OU Noise                    &$\sigma$             &$0.1$    \\
            \hline
        \end{tabular}
    \label{table: hyperparameters}
    \end{table}
    
    Our complete approach is described in Algorithm~\ref{ddpg}.

\section{RESULTS}
\subsection{Simulation Results}
We trained two different policies, Policy 1 (with action feedback) and Policy 2 (without action feedback) for 1500 episodes with randomly sampled targets from the manipulator work-space. All the hyper-parameters and simulation parameters are kept identical, with the feedback to the explored actions being the only distinction between the two policies. 

\begin{figure}[!ht]
\includegraphics[width=7.5cm, height=5cm]{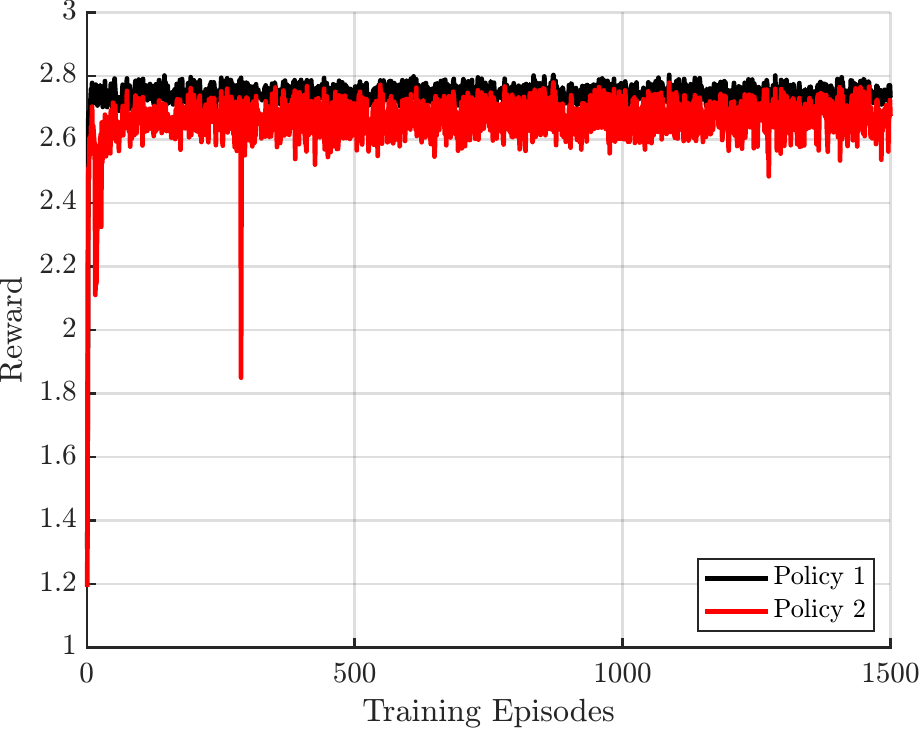}
\centering
\caption{Episode training rewards for Policy 1 (with feedback) and Policy 2 (without feedback) are shown. It is evident from the figure that Policy 1 is exploring efficiently because of the provided feedback. The feedback assists in selecting a meaningful action exploration.}
\vspace{-15pt}
\label{fig:episode_rewards}
\end{figure}

Fig~\ref{fig:episode_rewards} shows the cumulative reward for both the policies during the training episodes. Policy 1 constantly acquires better rewards than Policy 2 for each episode, validating our approach of efficient exploration using action feedback. 

\begin{figure}[!hbt]
\includegraphics[width=6cm, height=4.5cm]{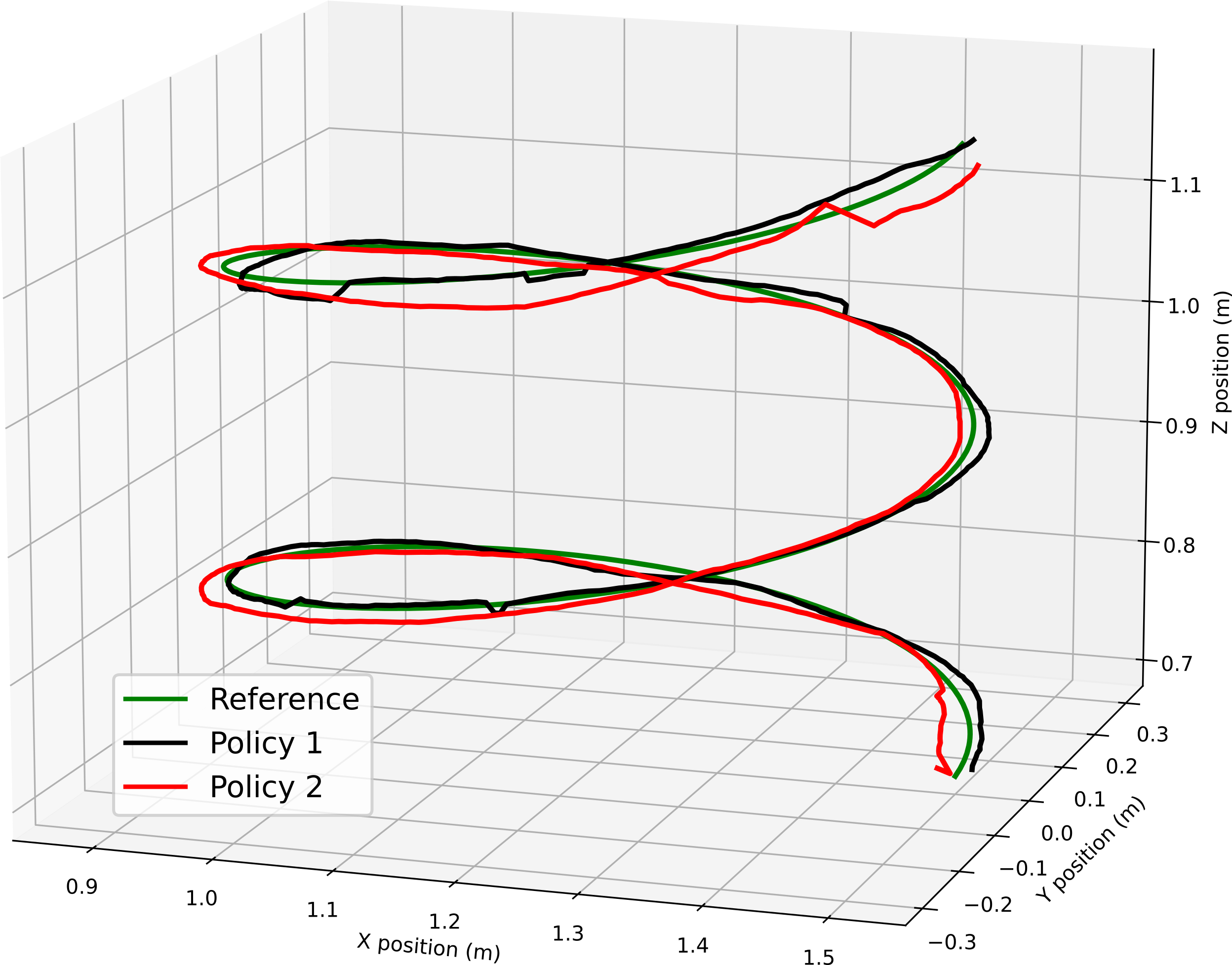}
\centering
\caption{In this figure we are comparing the 2 trained policies on a trajectory tracking task in simulation. Policy 1 performs better in tracking the helical trajectory in contrast to Policy 2.}
\label{fig:policy_comparison}
\end{figure}

\begin{figure}[!hbt]
    \includegraphics[width=7.5cm, height=5cm]{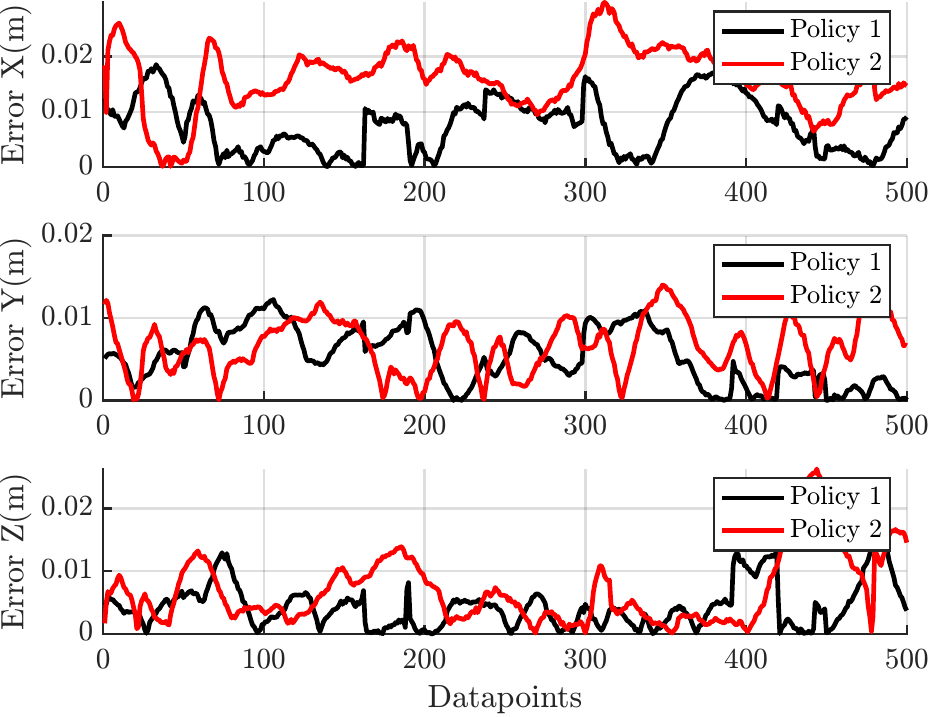}
    \centering
    \caption{The image shows the trajectory tracking error for Policy 1 and Policy 2.}
    \vspace{-15pt}
    \label{fig:policy_error}
\end{figure}

We validated both policies for tracking a helical trajectory. From the trajectory tracking results shown in Fig ~\ref{fig:policy_comparison}, it can be seen that the tracking accuracy for policy 1 is better than policy 2. The absolute tracking error is shown in ~\ref{fig:policy_error}. The Root Mean Squared Error (RMSE) for Policy 1 is [0.017, 0.008, 0.01], whereas for Policy 2 is [0.031, 0.017, 0.02]. Though we trained our RL agent within a defined actual manipulator work-space, we observed that the learned policy generalized the target reaching task and could perform trajectory tracking even outside the work-space on which it is trained.

\subsection{Real World Experiments}
We deployed Policy 1 directly on the real manipulator without any modifications to the outputs from the learned policy. We validated our learning-based controller approach in real-world experiments by tracking circular and helical trajectories. 

\begin{figure}[ht]
\includegraphics[width=7.5cm, height=6cm]{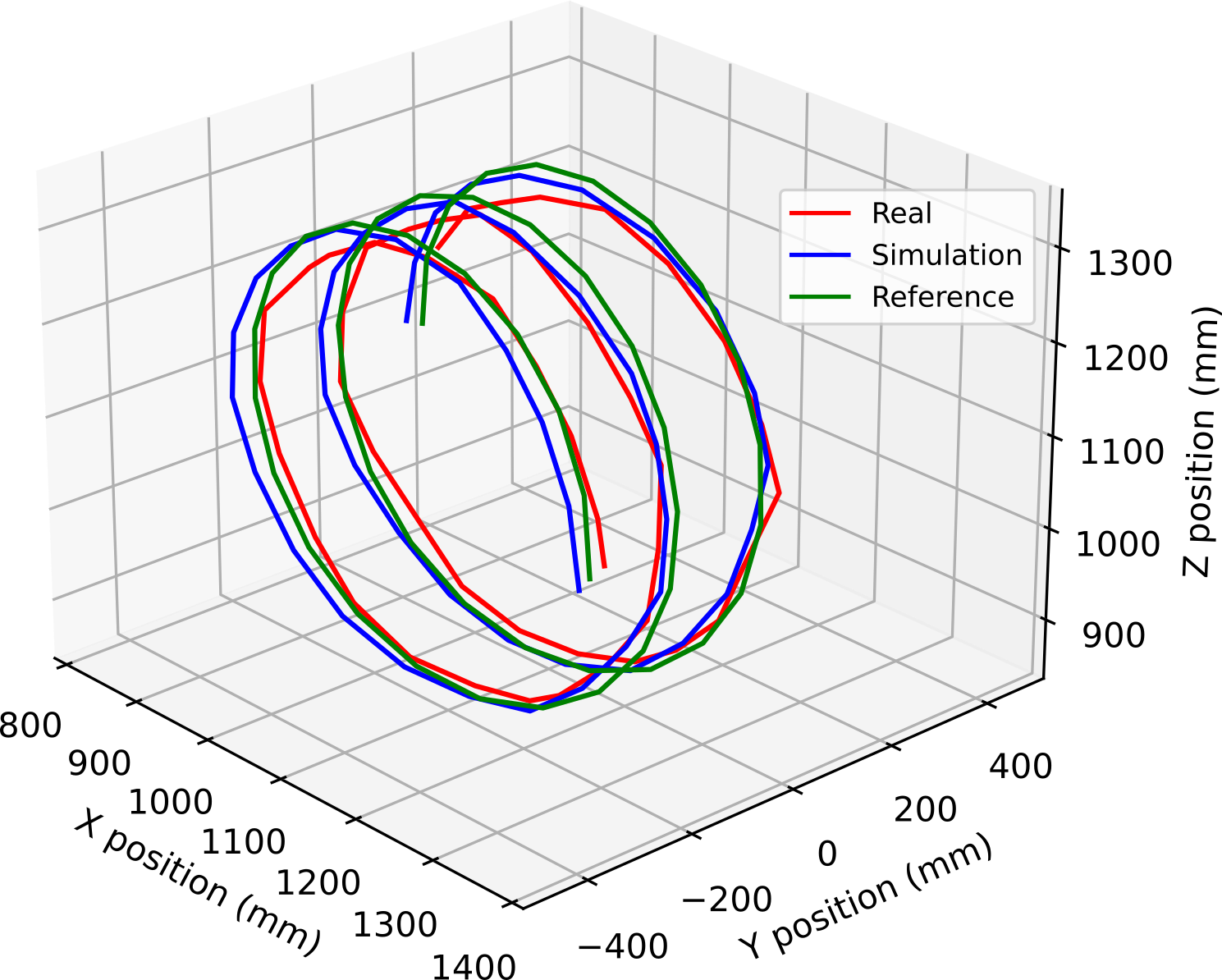}
\centering
\caption{The figure displays the real-world experiment results of trajectory tracking from deployed Policy 1 onto the manipulator. The tracking is performed well given the harsh dynamic conditions of our system. The results validate the Sim-2-Real transfer of our learning-based control approach.}
\vspace{-15pt}
\label{fig:real_traj}
\end{figure}

\begin{figure}[ht]
\includegraphics[width=7.5cm, height=5cm]{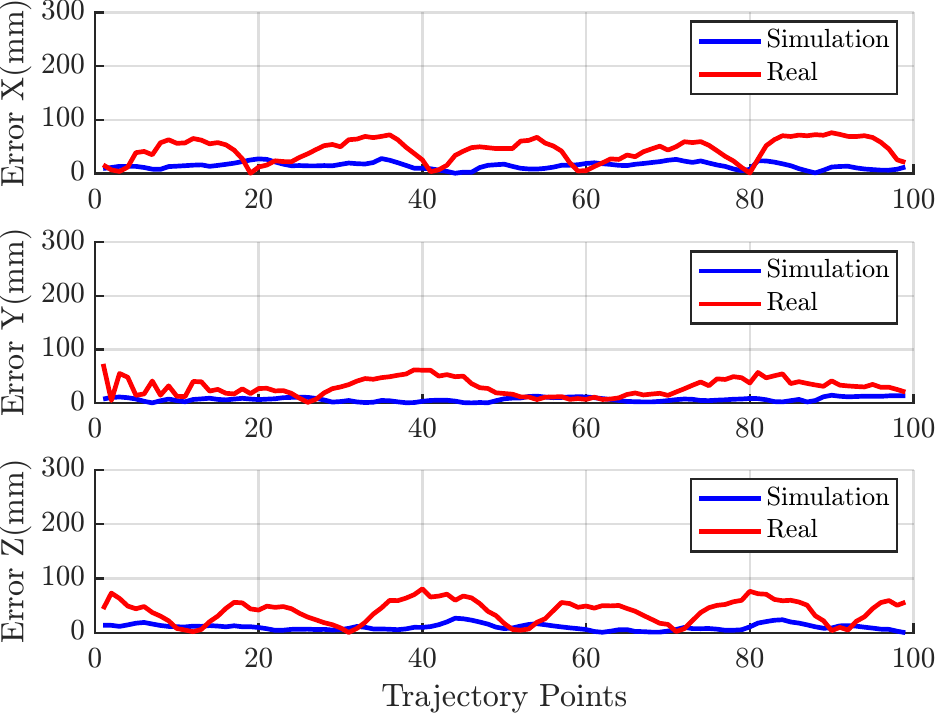}
\centering
\caption{The image visualizes the error in tracking of helical trajectory for real-world experiments. As foreseen for the same trajectory, the tracking error of the real experiment is bigger compared to simulation tracking error. However, the results are closely comparable. The periodic error is caused by the motion in the y-axis which causes the most sway motion of the gripper.}
\vspace{-15pt}
\label{fig:tracking_error}
\end{figure}

As we trained our control policy by detaching the gripper in simulation, we did not account for the dynamic sway of the gripper when the manipulator is in motion, which is currently out of the scope of our proposed approach. The real-world trajectory tracking experiments shows that the manipulator is successfully tracking the target trajectory, see Fig~\ref{fig:real_traj}, however, the unmodeled and unaccounted sway induced some tracking errors during the motion see ~\ref{fig:tracking_error}. 
Table ~\ref{table: traj_error} shows the  maximum tracking error for the helical trajectory in simulation and real-world using our learned controller.
    \begin{table}
    \centering
    \caption{Trajectory Tracking Errors}
        \begin{tabular}{|c|l|}
        \hline
            Experiment              &Max. Error (mm)  \\                   
            \hline
            \hline
            Simulation              &27.2, 14.5, 26.3   \\                
            \hline
            Real-World              &75.2, 80.1, 73.1     \\           
        \hline
        \end{tabular}
    \label{table: traj_error}
    \end{table}

    \begin{algorithm}[!htpb]
    \caption{DDPG with Action Feedback}\label{ddpg}
        \DontPrintSemicolon
                  \Kwinit{}
                    {\textit{Randomly initialize both actor  and critic  networks,}}\;
                     $\mu(s|\theta^\mu) \gets \theta^\mu$, $Q(s,a|\theta^Q)\gets \theta^Q$\; 
                    {\textit{Initialize target networks $Q^{'}$ and $\mu^{'}$,}}\;
                    $\theta^{Q^{'}} \gets \theta^{Q}$, $\theta^{\mu^{'}}  \gets \theta^{\mu^{}}$\;
                     {\textit{Initialize replay buffer}}\;
                  \Kwtrain{}
                    \For{\texttt{$n=1, n_{episodes}$}}{
                     {\textit{Reset environment}}\;
                     {\textit{Receive initial observation state $S$}}\;
                         \For{\texttt{$t=1, n_{steps}$}}{
                            \For{\texttt{$p=1, n_{actions}$}}{
                                    $a_p = \mu({s}_t|\theta^\mu)+\mathcal{N}_t$\;
                                \tcc*{according to current policy and exploration noise}
                                \Return $[a_{p_{1}},..., a_{p_{n}}]$
                            }
                            \KwFn:\;
                             \qquad return $\gets a_{p_{i}}, \vert min(x_{target} - x_{p_{i}})$\;
                             \textit{Set} $a_t$ $\gets$ $a_{p_{i}}$\;
                             \textit{Execute action:} $a_t$\;  
                             \textit{Observe:} reward $r_t$ and new state ${s}_{t+1}$\;
                             \textit{Store transition:} $({s}_t, a_t, r_t, {s}_{t+1})$ in $R$\;
                             \textit{Sample random batch:} from $n_{batch}$ transitions 
                             $({s}_i, a_i, r_i, {s}_{t+i})$ from $R$\;
                             \textit{Set:} $y_i = r_i + \gamma Q^{'} (s_{i+1}, \mu^{'} (s_{i+1} | \theta^{\mu^{'}})|\theta^{Q^{'}})$\;
                             \textit{Update critic by minimizing the loss:}\;
                             $L = \frac{1}{N} \sum_{i} (y_i - Q (s_i,a_i|\theta^{Q}))^2$\;
                             \textit{Update actor policy using sampled policy\;
                             gradient:}\;
                             $\nabla_{\theta^{\mu}} J 
                             \approx \frac{1}{N} \sum_i \nabla_a Q (s,a|\theta^{Q}) |_{s = s_i, a=\mu(s_i)} \nabla_{\theta^{\mu}} \mu(s|\theta^\mu) |_{s_i} $\;
                             \;
                             \textit{Update target networks,}\;
                             $\theta^{Q^{'}} \gets \tau \theta^{Q} + (1 - \tau)\theta^{Q^{'}}$\;
                             $\theta^{\mu^{'}} \gets \tau \theta^{\mu} + (1 - \tau)\theta^{\mu^{'}}$
                             }
                    }
    \end{algorithm}

\section{CONCLUSIONS AND FUTURE WORK}
The presented results demonstrate the direct application of RL to heavy-duty manipulators and the feasibility of directly deploying a control policy entirely learned in simulation to physical forestry cranes. The main advantage of the presented approach is that no mathematical formulation either of kinematics or dynamics is required. For our approach, we do not need the geometry information to acquire the cylinder-joint mapping which is required for the automation of such manipulators. Our approach inherently adapts the actuation dynamics which in general is a complex problem involving numerous external factors. Our controller requires minimal system information which can be easily acquired by retrofitting the manipulator, thus making the automation of such heavy manipulators very efficient and economical. We made use of the available information in a simple and elegant approach to make the controller learning process much more efficient by providing feedback on the exploration actions and choosing the best one among the action candidates. However, our real-world experiment results suffered from tracking errors, mainly due intervening dynamic factors (gripper sway, backlash, actuation inaccuracies) and poorly tuned low-level control. In contrast to these intervening factors and given the fact that our controller is trained only on 1500 data points sampled from the complete manipulator trajectory, the tracking accuracy is remarkable. The controller performance can be greatly improved, by training on more data points and providing a finely tuned low-level controller.

To address the problem of gripper sway, in future work we will extend our framework to integrate the sway motion during the learning process to model a compensating or aggressive control policy. We also plan to incorporate a generalized Long Short Term Memory (LSTM) based backlash model, to also take the backlash motion into account during training. Even though our feedback model facilitates the controller in an efficient exploration and learning, it still contains minor inaccuracies which might be affecting the learning process. A better feedback model will undoubtedly improve the controller performance. For more complex manipulation tasks we plan to use curriculum learning\cite{bengio2009curriculum}, which has been shown to accelerate and improve the learning process.


\bibliographystyle{IEEEtran}

\bibliography{bibtex/literatur}

\end{document}